# Deep Learning-Based UAV Aerial Triangulation without Image Control Points

Jiageng Zhong[1], Ming Li[1], Jiangying Qin[1], Hanqi Zhang[1]

[1]State Key Laboratory of Information Engineering in Surveying Mapping and Remote Sensing,
Wuhan University, Wuhan 430079 China,
Email: zhongjiageng@whu.edu.cn, lisouming@whu.edu.cn, jy_qin@whu.edu.cn, hqzhang@whu.edu.cn

**ABSTRACT:** The emerging drone aerial survey has the advantages of low cost, high efficiency, and flexible use. However, UAVs are often equipped with cheap POS systems and non-measurement cameras, and their flight attitudes are easily affected. How to realize the large-scale mapping of UAV image-free control supported by POS faces many technical problems. The most basic and important core technology is how to accurately realize the absolute orientation of images through advanced aerial triangulation technology. In traditional aerial triangulation, image matching algorithms are constrained to varying degrees by preset prior knowledge. In recent years, deep learning has developed rapidly in the field of photogrammetric computer vision. It has surpassed the performance of traditional handcrafted features in many aspects. It has shown stronger stability in image-based navigation and positioning tasks, especially it has better resistance to unfavorable factors such as blur, illumination changes, and geometric distortion. Based on the introduction of the key technologies of aerial triangulation without image control points, this paper proposes a new drone image registration method based on deep learning image features to solve the problem of high mismatch rate in traditional methods. It adopts SuperPoint as the feature detector, uses the superior generalization performance of CNN to extract precise feature points from the UAV image, thereby achieving high-precision aerial triangulation. Experimental results show that under the same pre-processing and post-processing conditions, compared with the traditional method based on the SIFT algorithm, this method achieves suitable precision more efficiently, which can meet the requirements of UAV aerial triangulation without image control points in large-scale surveys.

## 1. INTODUCTION

The UAV-based aerial triangulation using POS (position and orientation system) is extremely rapid and can easily cover the survey area. The POS provides the measurement of the position and orientation of the camera so that each image and pixel can be georeferenced to the Earth without the need for image control points, and the most important and most commonly used data is the position data collected from Global Navigation Satellite Systems (GNSS). Although the drone aerial survey has the advantages of low cost and high efficiency, it is still a problem to achieve high accuracy. One important factor that would affect global accuracy is the precision of feature matching which directly influences the precision of the entire registration. Therefore, accurate features extraction is the basic and key technique in aerial triangulation.

The feature extraction consists of keypoint detection and description and has been used in computer vision task for a long time. The keypoint or feature can be described as a specific meaningful structure, but it is not clear what are the relevant keypoints for an arbitrary input image (Mukherjee and et al., 2015). The function of a feature detector is to detect keypoints and their corresponding descriptors. In the past decades, feature detectors have been an activate area of research. Among the many detectors, SIFT (Scale-Invariant Feature Transform) (Lowe, 2004) is the most representative and influential one. SIFT aims to solve the image rotation, affine transformations, intensity, and viewpoint change in matching features (Karami and et al., 2017). It generally includes two major steps. It firstly convolves the image with Gaussian filters at various scales and finds scale invariant keypoints via estimating a scale space extreme. Then, for each keypoint, the local image descriptor is computed based on image gradient magnitude and orientation (Lowe, 2004). And there are many other SIFT-like detectors such as SURF (Speed up Robust Feature) (Bay and et al., 2008) and ORB (Oriented FAST and Rotated BRIEF) (Rublee and et al., 2011) which are more efficient than SIFT.

With the rapid development of deep learning methods and the increasing demand for better feature detection, new feature detectors emerged, most of which are based on convolutional neural network. Different from the classical algorithms, deep learning approaches can learn abstract image features from high-dimensional data in an end-to-end fashion instead of relying on handcrafted features such as distinctive corners (Zeiler and Fergus, 2014). As convolutional neural networks learn features based on supervision, their performance heavily relies on ground truth information (Bojanić and et al.,2019). In other word, the key is usually a large dataset of 2D ground truth locations labeled by human annotators. Unlike these approaches, a novel detector named SuperPoint (DeTone and et al., 2018) is supervised by itself and works well for matching tasks.

In this paper, a new aerial survey method without image control points is proposed, and SuperPoint is applied to replace traditional feature detector in the aerial triangulation flow. Through the multiple perspectives experiment, it is shown that the new method is able to achieve suitable precision more efficiently, compared to those based on classic feature detectors.

## 2. METHODOLOGY

### 2.1 Big Picture

Referring COLMAP's incremental Structure-from-Motion pipeline (Schonberger and Frahm, 2016), our method can be divided into three stages as shown in Figure 1. The first stage is to prepare UAV images and corresponding position data which contains latitude and longitude location information. Note that the position data should be converted in Gauss-Kruger Projection.

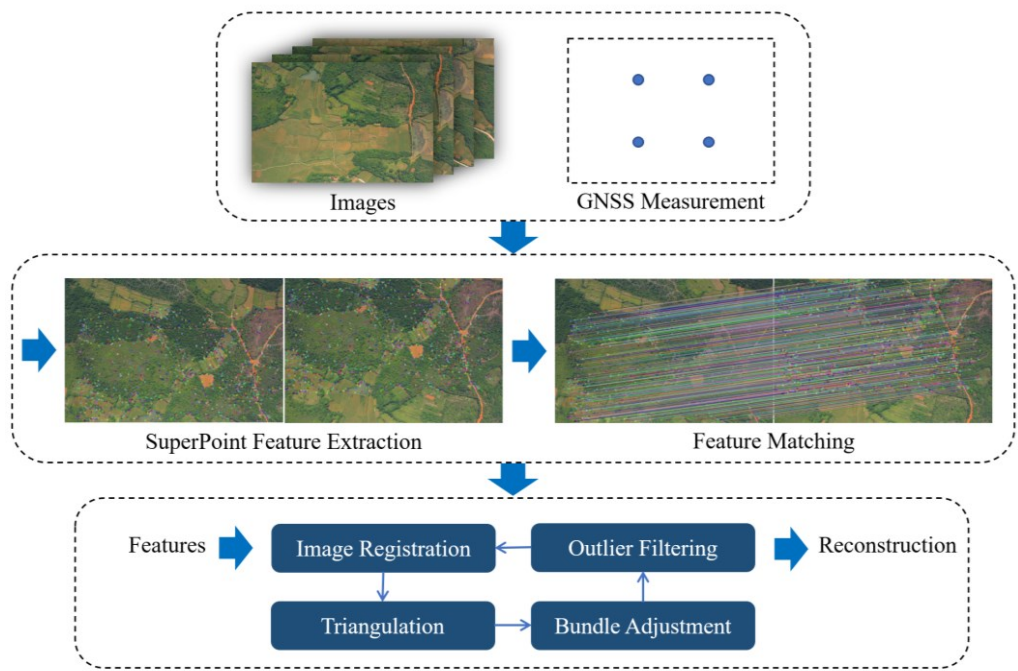

Figure 1. A flow chart of our aerial survey method

The second stage is correspondence search which finds overlap in the input images and identifies projections of the same points in overlapping images. For each image, the first step is to extract features which are invariant under radiometric and geometric changes. In traditional aerial triangulation, SIFT (Lowe, 2004) is applied mostly, here it is replaced by SuperPoint which can also output L2-normalized fixed length descriptors. As for feature matching, a matcher that combines Lowe's ratio test matcher (Lowe, 2004) and Nearest Neighbor search is adopted to improve the matching accuracy. Based on matched features, images that cover the same scene part are discovered. So the output of the second stage is a set of potentially overlapping image pairs and their associated feature correspondences (Schonberger and Frahm, 2016). In addition, there are typically geometric verification that uses projective geometry to verify the matches.

The third stage is mainly carried out in four steps. Based on the output of the second stage, new images can be registered. Next, as newly registered images lead to increase scene coverage, new scene points can be triangulated and added to the scene structure. Then, considering the possible problem of error accumulation in reconstruction, BA (Bundle Adjustment) (Triggs and et al., 1999) is applied to refine camera parameters and point parameters via minimizing the reprojection error. Finally, the outliers are filtered. This iterative strategy can significantly improve completeness and accuracy (Schonberger and Frahm, 2016).

Through the workflows above, the aerial survey without image control points is finished. Specifically, all the images are aligned and a set of sparse point cloud of the survey area is formed.

## 2.2 SuperPoint Network

SuperPoint is an encoder-decoder architecture and is a fully-convolutional neural network which operates on a full-sized image. Its structure is shown in Figure 2. Its shared encoder processes the input image at first and branches into two decoders, one for interest point detection and the other for interest point description. This strategy is quite different from traditional system which first detects keypoints and then calculates the descriptors.

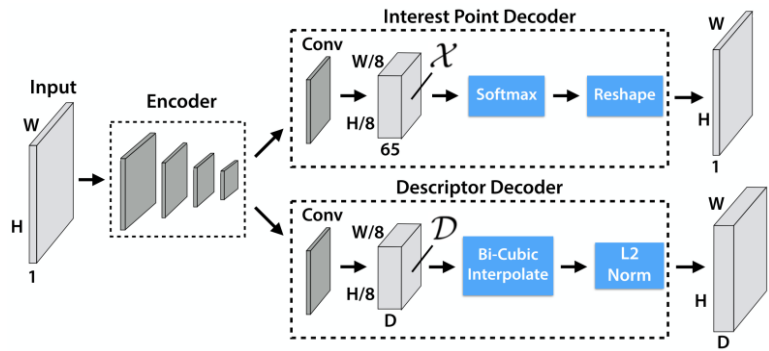

Figure 2. Structure of SuperPoint Network (DeTone and et al., 2018)

It should be noted that Superpoint adopts self-supervised training strategy. It firstly trains a base detector called MagicPoint based on supervision from a synthetic dataset, where keypoints can be determined unambiguously. Then the detector capacity is expanded to real images using homographic adaption. Finally, a keypoint descriptor is computed by an additional subnetwork.

## 3. EXPERIMENTS

### 3.1 Data Preparation

In this section, we present experiment results of the traditional method (based on SIFT) and our method for comparison. All experiments in this paper are based on two datasets collected from Chongqing. Each dataset contains a UAV image sequence and corresponding POS data of a scene. The GSD (ground sample distance) of the aerial images is 0.2 m, and the GNSS standard horizontal and vertical precision is 1 cm and 3 cm respectively. In addition, there are also ground known points for precision check. To be more specific, for Scene 1 and Scene 2, there are 24 images and 6 images respectively. The heading overlap and side overlap rate were set to 80% and 60% respectively. These can meet the specification of topographic mapping at small scale.

### 3.2 System Runtime

The run-time of SuperPoint and SIFT is measured using a RTX 2060 GPU. The SuperPoint architecture is implemented with Pytorch deep learning library (Paszke and et al., 2019). The average run-time of different algorithms is measured as shown in Table 1. As the inference of the deep model is done in a single forward propagation step, the run-time of a single forward pass is measured to be about 148 ms. And SIFT takes about 368 ms to process one image. It can be seen that SuperPoint executes more efficiently than SIFT and may be applied in real time surveying.

Table 1.  Mean execution times

| Algorithm | SuperPoint | SIFT |
|---|---|---|
| Run-time | 148 ms | 368 ms |

### 3.3 Feature Extraction and Matching

Several comparative experiments are carried out for qualitative and quantitative evaluation of the performance of the keypoint detector and descriptor generator on the datasets.

The appearance and distribution of keypoints from different detectors are intuitively demonstrated in Figure 3, and the image is from the dataset of Scene 1. It can be observed that SuperPoint produces fewer feature points compared to SIFT, and its location distribution is more dispersive. For instance, there is a tract of farmland in the top left of images, SIFT can hardly extract keypoints, and SuperPoint can extract more well-distributed keypoints. Instead, for tree or residential areas that have rich texture information, there are more dense points produced by SIFT.

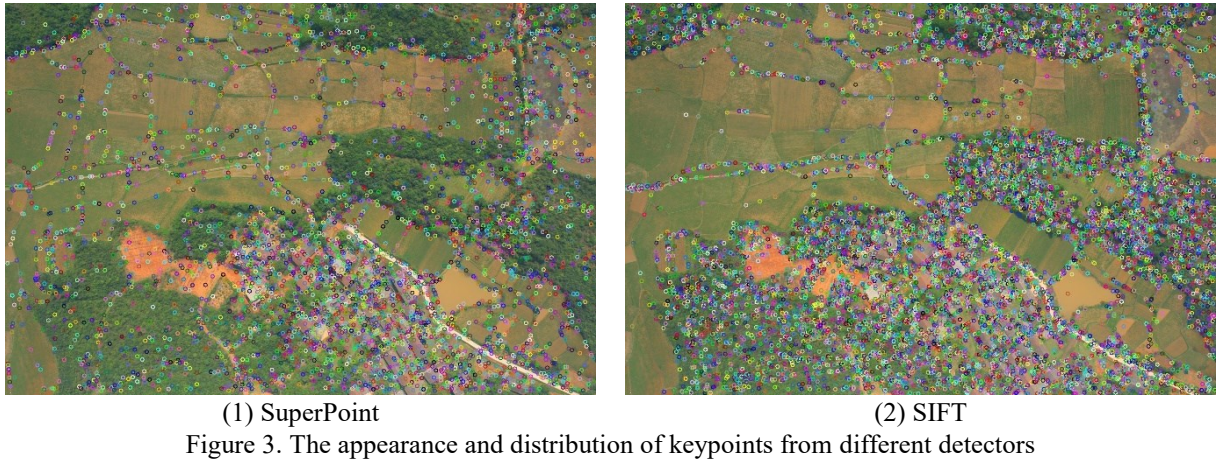

(1) SuperPoint                                    (2) SIFT

Figure 3. The appearance and distribution of keypoints from different detectors

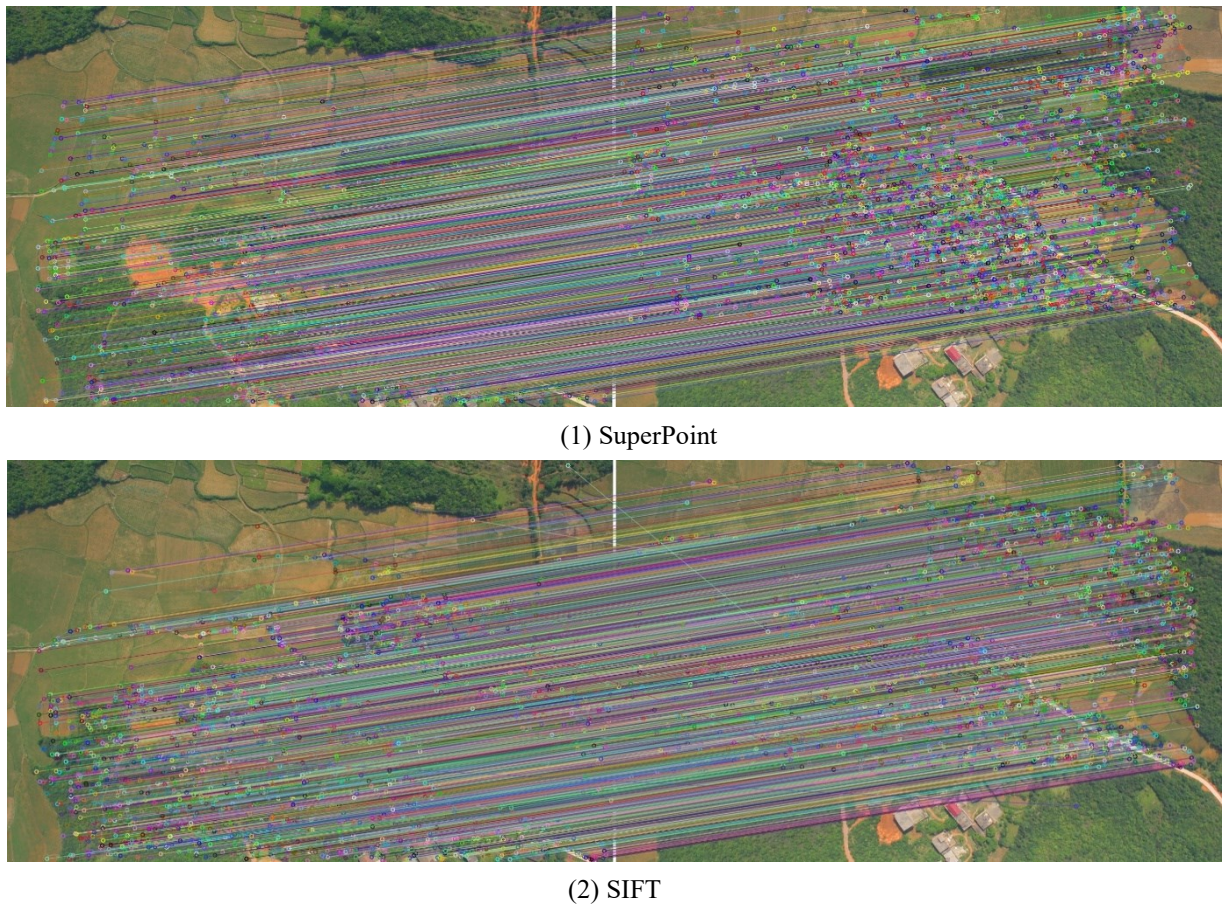

(1) SuperPoint

(2) SIFT

Figure 4. Qualitative result of matching

Then, to evaluate the performance of the descriptors, the extracted features are matched by a matcher that combines Lowe's ratio test matcher (Lowe, 2004) and Nearest Neighbor search. The ratio test checks if matches are ambiguous and should be removed, because the probability that a match is correct can be determined by taking the ratio of distance from the closest neighbor to the distance of the second closest (Lowe, 2004). A qualitative example of SuperPoint versus SIFT is shown in Figure 4 and the distance ratio is set to 0.7. SuperPoint tends to produce a larger number of correct matches which densely cover the image while there are several mismatches in the result of SIFT.

The statistics analysis of the quality of descriptors is performed. Figure 5 shows the match rates and mismatch rates under different distance ratios for real image data. The match rate is defined as the ratio of matched keypoints to all keypoints, and the mismatch rate is defined as the ratio of keypoints which are matched falsely to matched keypoints.

Figure 5(a) and 5(b) shows that SuperPoint has higher match rates and lower mismatch rates in most cases. For SIFT detector, there are too many mismatches to estimate the pose when the distance ratio is greater than 0.8. The ratio is typically set between 0.5 to 0.8 in application, and in this range, SuperPoint can achieve zero mismatch. Therefore, it is reasonable to suppose that SuperPoint is able to produce better descriptors.

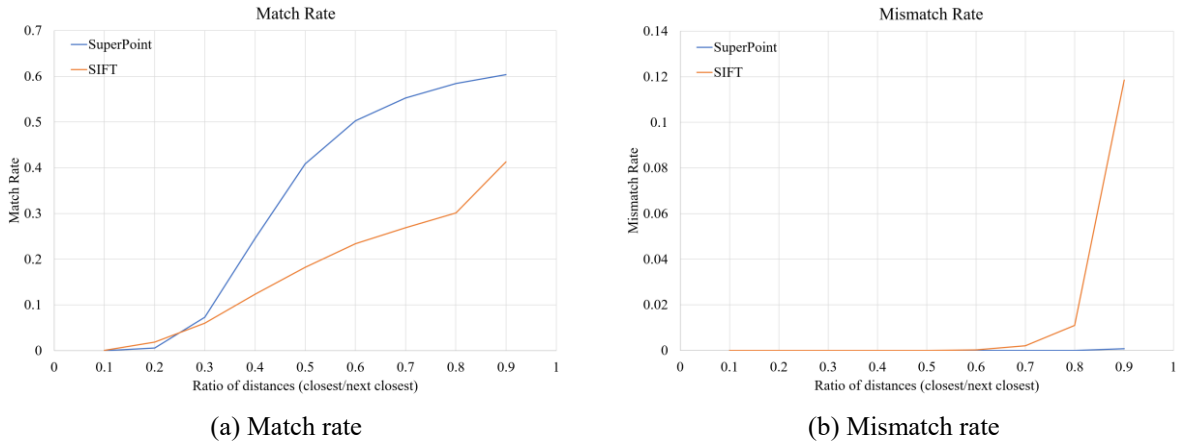

(a) Match rate                           (b) Mismatch rate

Figure 5. The statistical results of feature matching

Table 2. Relative orientation error

| Algorithm | SuperPoint | SIFT |
|---|---|---|
| Reprojection Error | 0.1454 pixel | 0.1008 pixel |

A relative orientation process recreates relative translation and angular relationships between two successive overlapping images (Tjahjadi and Agustina, 2019). In this paper, the reprojection error of relative orientation is used as the metric for evaluating the quality of keypoints. Table 2 displays the errors of the image pair in Figure 4. SIFT performs better on this metric as SuperPoint has a higher reprojection error. This is likely due to the fact that SIFT performs extra sub-pixel localization, while SuperPoint does not perform this step.

### 3.4 Aerial Triangulation

Using our new aerial survey method described in Section 2, the aerial triangulation without image control points is carried out on datasets of Scene 1 and Scene 2. The reprojection errors in bundle adjustment are displayed in Table 3, and the camera position errors are displayed in Table 4. Due to extra sub-pixel localization, the SIFT-based method reaches higher precision. As for camera position, our method has slightly higher precision, presumably because the keypoints extracted by SuperPoint distribute more evenly.

Table 3. Reprojection errors in bundle adjustment

|  | Our Method | SIFT |
|---|---|---|
| Scene 1 | 0.387 pixel | 0.332 pixel |
| Scene 2 | 0.412 pixel | 0.353 pixel |

Table 4. Camera position errors

|  | ERROR | Our Method | SIFT |
|---|---|---|---|
| Scene 1 | X error | 0.603 m | 0.530 m |
|  | Y error | 0.716 m | 1.004 m |
|  | Z error | 0.202 m | 0.166 m |
|  | XY error | 0.936 m | 1.136 m |
|  | XYZ error | 0.958 m | 1.148 m |
| Scene 2 | X error | 0.451 m | 0.425 m |
|  | Y error | 0.570 m | 0.422 m |
|  | Z error | 0.791 m | 1.043 m |
|  | XY error | 0.727 m | 0.599 m |
|  | XYZ error | 1.074 m | 1.203 m |

Table 5. Error of the checkpoint

| ERROR | Our Method | SIFT |
|-------|-----------|------|
| X error | -1.821 m | -2.444 m |
| Y error | -2.217 m | -1.635 m |
| Z error | -3.755 m | -3.925 m |
| XY error | 2.870 m | 2.940 m |
| XYZ error | 4.726 m | 4.904 m |

A known point in Scene 2 is used as the checkpoint for precision check, and Table 5 displays the checkpoint error. The comparison result is consistent with Table 4. The results of experiments illustrate that our method is likely to reach higher precision compared to traditional SIFT-based method, which confirms that learned representations for descriptor matching outperform hand-tuned representations.

## 4. CONCLUSION

This paper presents a new aerial survey method without image control points, which adopts SuperPoint as feature detector. A series of comparative experiments illustrate that our method has obvious advantage in efficiency, keypoint distribution and matching quality, and it can achieve suitable precision. So, it can be concluded that our method is capable to meet the application requirements of aerial triangulation. Future work will comprehensively evaluate the performance of our method with more experiment.

This paper has shown that the deep learning method outperforms traditional methods in many aspects, therefore, we can consider that deep learning-based aerial survey would have an expected future.

## ACKNOWLEDGEMENTS

This research was funded by the National Key R&D Program of China, grant numbers 2018YFB0505400, the National Natural Science Foundation of China (NSFC), grant number 41901407 and the LIESMARS Special Research Funding.